\documentclass{article}

\PassOptionsToPackage{numbers, compress}{natbib}
\usepackage[preprint]{neurips_2025}

\usepackage[utf8]{inputenc} % allow utf-8 input
\usepackage[T1]{fontenc}    % use 8-bit T1 fonts
\usepackage{hyperref}       % hyperlinks
\usepackage{url}            % simple URL typesetting
\usepackage{booktabs}       % professional-quality tables
\usepackage{amsfonts}       % blackboard math symbols
\usepackage{nicefrac}       % compact symbols for 1/2, etc.
\usepackage{microtype}      % microtypography
\usepackage{xcolor}         % colors
\usepackage{multirow}
\usepackage{graphicx}
\usepackage{float}
\usepackage{amssymb}
\usepackage{subfigure}
\newcommand{\name}{\textsc{LLM4-IC8K}}
\newcommand{\namedata}{\textsc{ICGeo8K}}
\newcommand{\namedatagenerated}{\textsc{ICGeo8K-Syn}}
\newcommand{\nametest}{\textsc{ICGeoQA}}

\newcommand{\ylq}{\textcolor{green}}

\newcommand{\ie}{{\em i.e.}}

\title{A Large Language Model Powered Integrated Circuit Footprint Geometry Understanding}

\author{%
  Yida Wang\textsuperscript{1}\thanks{Both authors contributed equally.}\quad
  Taiting Lu\textsuperscript{2}\footnotemark[1]\quad
  Runze Liu\textsuperscript{2}\quad
  Lanqing Yang\textsuperscript{1}\thanks{Corresponding author.}\quad 
  Yifan Yang\textsuperscript{3}\thanks{Corresponding author.}\quad\\
  \textbf{
  Zhe Chen\textsuperscript{1}\quad
  Yuehai Wang\textsuperscript{1}\quad
  Yixin Liu\textsuperscript{1}\quad
  Kaiyuan Lin\textsuperscript{2}\quad
  Xiaomeng Chen\textsuperscript{1}\quad
  }\\[6pt]
  \textbf{
  Dian Ding\textsuperscript{1}\quad
  Yijie Li\textsuperscript{4}\quad
  Yi\,-Chao~Chen\textsuperscript{1}\quad
  Yincheng Jin\textsuperscript{5}\quad
  Mahanth Gowda\textsuperscript{2}
  }\\[6pt]
  \textsuperscript{1}Shanghai~Jiao~Tong~University\quad
  \textsuperscript{2}Pennsylvania State University\\
  \textsuperscript{3}Microsoft Research\quad
  \textsuperscript{4}National University of Singapore\quad
  \textsuperscript{5}Binghamton University\\[4pt]
  \texttt{\{yidawang, yanglanqing, chenzhe2003, yuehaiw, panh09,} \\ 
  \texttt{dingdian94, yichao, sjtu\_chenxm\}@sjtu.edu.cn}\\
  \texttt{\{txl5518, rml6043, kjl6302, mahanth.gowda\}@psu.edu}\quad\\
  \texttt{yifanyang@microsoft.com}\quad
  \texttt{louise\_l@alumni.sjtu.edu.cn}\quad\\
  \texttt{yijieli@nus.edu.sg}\quad
  \texttt{yjin5@binghamton.edu}
}

\begin{document}

\maketitle

\begin{abstract}
  % The abstract paragraph should be indented \nicefrac{1}{2}~inch (3~picas) on
  % both the left- and right-hand margins. Use 10~point type, with a vertical
  % spacing (leading) of 11~points.  The word \textbf{Abstract} must be centered,
  % bold, and in point size 12. Two line spaces precede the abstract. The abstract
  % must be limited to one paragraph.
 Printed-Circuit-board (PCB) footprint geometry labeling of integrated circuits (IC) is essential in defining the physical interface between components and the PCB layout, requiring exceptional visual perception proficiency. However, due to the unstructured footprint drawing and abstract diagram annotations, automated parsing and accurate footprint geometry modeling remain highly challenging. Despite its importance, no methods currently exist for automated package geometry labeling directly from IC mechanical drawings. In this paper, we first investigate the visual perception performance of Large Multimodal Models (LMMs) when solving IC footprint geometry understanding. Our findings reveal that current LMMs severely suffer from inaccurate geometric perception, which hinders their performance in solving the footprint geometry labeling problem. To address these limitations, we propose \textbf{{\name}}, a novel framework that treats IC mechanical drawings as images and leverages LLMs for structured geometric interpretation. To mimic the step-by-step reasoning approach used by human engineers, {\name} addresses three sub-tasks: perceiving the number of pins, computing the center coordinates of each pin, and estimating the dimensions of individual pins.
%, where each "pin" refers to a pad or terminal on the IC package used for electrical connection, and each coordinate represents its precise location relative to the IC origin in the PCB layout. 
We present a two-stage framework that first trains LMMs on synthetically generated IC footprint diagrams to learn fundamental geometric reasoning and then fine-tunes them on real-world datasheet drawings to enhance robustness and accuracy in practical scenarios.
%Our method achieves 71.6\% $IoU_{IC}$ accuracy, comparable to traditional EDA tools, while reducing design time to 0.26 minutes per sample, a 58× speedup over manual creation. 
To support this, we introduce {\namedata}, a multi-modal dataset with 8,608 labeled samples, including 4138 hand-crafted IC footprint samples and 4470 synthetically generated samples. Extensive experiments demonstrate that our model outperforms state-of-the-art LMMs on the proposed benchmark. The accurate translation of footprint diagrams enabled by {\name} contributes to advancing standardization within the PCB industry. 

  % We propose a two-stage training pipeline, where a large language model initially learns basic geometric understanding from synthetic diagrams and then adapts to complex real-world datasheet images, effectively enabling automated and accurate IC footprint geometry labeling.
  % To support model training and evaluation, we also introduce a new dataset of 4,138 hand-crafted annotated IC mechanical drawings with ground-truth pin-level geometry labels.
  % Experimental results demonstrate that
  % % {\namedata} delivers \ylq{XXX: detail} significant metric over existing LLM models in IC package geometry labeling, highlighting its generalizability.
  % {\name} achieves the highest accuracy in IC geometry understanding tasks, with an overall $IoU_{IC}$ of $71.6\%$,  over existing general LLM models in IC package geometry labeling.

\end{abstract}

\section{Introduction}
\label{sec:intro}

% Motivation:

% Introduce IC geometry labeling.

% The importance of this application.

% The difficulty of this application.

% Current models fail to carry out this application.

% main contributions

%
Large Language Models (LLMs) have shown promising performance in geometry and spatial reasoning \cite{gao2023gllavasolvinggeometricproblem}\cite{cheng2024spatialrgptgroundedspatialreasoning}, proof and logic reasoning \cite{NEURIPS2023_10803064}, and Graph and Set Theory \cite{wang2024llms}.
However, existing researches mainly focus on abstract geometric reasoning within textual and synthetic visual domains, whereas complex, real-world engineering tasks, such as understanding footprint geometry \ylq{of} Printed-Circuit-board (PCB) and integrated circuit (IC) package drawings, remain underexplored.
Since general LLMs are not designed to handle dense IC footprint diagrams that include many annotations and many engineering drawing labels (as shown in Figure~\ref{fig:label_example} and Appendix~\ref{ssec: sample_example}), LLMs struggle to perform accurately on such visually complex tasks. 
Furthermore, collecting trainable diagram-description pairs for LLM involves massive labor efforts and expert experiences, and there are no publicly available datasets containing IC diagrams that can be directly utilized for training LLMs in IC geometry understanding.

\begin{figure}
    \centering
    \includegraphics[width=1\linewidth]{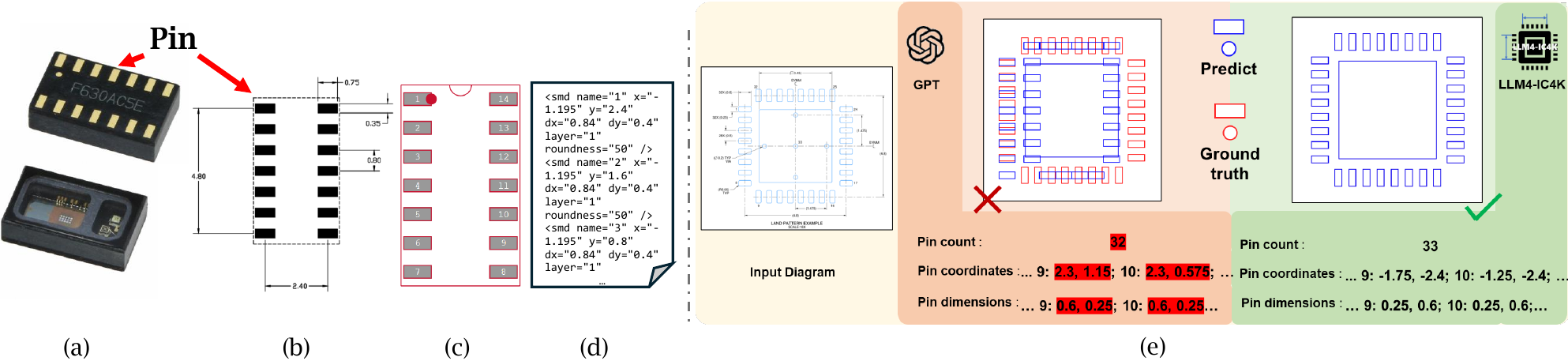}
        \vspace{-0.2in}
		\caption{Illustration of IC geometry understanding problem. 
            % (a) Example IC packages showing visible pins that serve as the physical electrical and mechanical interfaces of the chip.
            % (b) Corresponding footprint diagram from the IC datasheet, which provides geometric specifications, such as pin positions and dimensions, for designing the land pattern.
            % (c) Visualization of EDA source file, based on the footprint diagram from the datasheet.
            % (d) EDA footprint source file.
            (a) Example IC packages with visible pins, serving as electrical and mechanical interfaces.
            (b) Corresponding datasheet footprint diagram showing pin positions and dimensions for PCB design.
            (c) Visualization of EDA footprint based on the datasheet diagram.
            (d) Raw EDA footprint source file.
            (e) Visualization of an IC footprint parsing task: GPT and {\name} are given the same input diagram and tasked with extracting pin count, pin center coordinates, and pin dimensions. GPT produces incorrect answers, with errors highlighted in \textcolor{red}{red}, such as miscounted pin numbers or incorrect coordinate values. 
            % In contrast, {\name} accurately identifies all geometric attributes, demonstrating stronger spatial and numerical reasoning. 
            }
		\label{fig:banner}
        \vspace{-0.3in}
\end{figure}

PCB footprint geometry understanding has significant value in industrial production and PCB design because it ensures accurate component placement and reliable electrical connections on the PCB. 
In the PCB footprint geometry understanding, pins of the footprint define where and how the IC connects to the PCB board, determining both the physical placement and the electrical pathways needed for the circuit to function properly (as depicted in Figure \ref{fig:banner}).
Inaccuracies in pin size or placement can lead to mismatched impedance or unintended parasitic effects. 
These issues can significantly degrade the performance and reliability of high-speed or sensitive circuits \cite{sierra2023parasitic, intel2023impedance}.
In industrial production, a precise footprint ensures accurate component placement by automated assembly machines. 
Misaligned or incorrect footprint causes soldering defects, such as tomb-stoning or short circuits, resulting in production delays, and higher costs \cite{altium2022footprint}.
There have been numerous industrial electronic design automation (EDA), such as IPC compliant footprint wizard by Altium \cite{altiumIPCWizard}, Package Generator by Autodesk EAGLE \cite{autodeskEagleLibrary2020}, and Footprint Editor by KiCad \cite{KiCadFootprint}, etc.
While these tools provide manual footprint creation, they often require significant user input and lack automation in interpreting datasheet drawings. 
Motivated by the promising capabilities of LLMs in geometric and spatial reasoning, we first investigate their visual interpretation abilities to automate the perception of IC footprint information from datasheet drawings.
% Additionally, not all integrated circuits (IC) strictly adhere to IPC-7351 standards \cite{ultralibrarianIPC7351} or uniform datasheet drawing formats, causing manual adjustments to accommodate unique package geometries.

Understanding a precise IC footprint involves three core tasks: 
% (1) It requires identifying the number of each pin to ensure correct electrical connectivity. 
(1) Identifying \textbf{the number of pins} correctly to ensure accurate electrical connectivity.
% (2) The exact dimensions of each pin must be determined to support reliable soldering and mechanical integrity.
(2) Accurately determining \textbf{the spatial arrangement of all pins} relative to a clearly defined reference point (origin), such as the component's center or a specific pin, to ensure correct placement and alignment of the component on the PCB.
(3) Determining \textbf{the exact size of each pin} is important to make sure the electronic parts are firmly attached and work properly. 
If the pin sizes in a PCB design are wrong, the electronic parts might not fit properly or connect correctly, resulting in weak electrical connections and short circuits.
% This can lead to poor electrical contact, short circuits, or the parts not working at all, which causes the whole board to fail or malfunction.
% (3) The spatial layout of all pins must be accurately computed relative to a defined origin to facilitate proper component placement on the PCB.
% However, most existing LLMs struggle to understand IC footprint geometry because they lack the spatial reasoning and domain-specific visual-textual alignment needed to accurately interpret technical drawings and translate them into precise geometric layouts. 
However, existing large language models typically struggle to interpret IC footprint geometry accurately. They lack the spatial reasoning capabilities and the domain-specific alignment of visual and textual information necessary to reliably comprehend technical drawings and translate them into precise geometric layouts.

In this paper, we propose {\name} (\underline{\textbf{L}}arge \underline{\textbf{L}}anguage \underline{\textbf{M}}odel for (\underline{\textbf{4}}) \underline{\textbf{I}}ntegrated \underline{\textbf{C}}ircuit - 8K), a two-stage training framework designed to enable large language models to accurately understand IC footprint geometry and automate the footprint generation process.
by mimicking the step-by-step reasoning approach used by human engineers.
% In the first stage, we use a synthetically generated dataset with clean, structured diagrams to teach the model basic geometric reasoning, such as identifying pin count, positions, and dimensions through multi-round dialogue.
In the first stage, we employ a synthetically generated dataset comprising clean and structured diagrams to address the challenges of collecting and annotating real-world training data, which is both labor-intensive and requires domain expertise.
% Synthetic data also enables the efficient creation of diverse examples across various IC package types.
% In the second stage, the model is further trained on real-world footprint drawing from datasheets containing diverse and complex layout diagrams, helping it generalize to noisy annotations and varied visual styles.
In the second stage, the model is further trained on real-world footprint drawings from datasheets (shown in Appendix \ref{ssec: sample_example}), which contain dense annotations and varied geometric labeling. 
This stage is essential for adapting the model to the complexity and noise present in practical applications, enabling it to generalize effectively to diverse visual styles and layout conventions encountered in real-world IC design.
To further improve performance of the model, we structure the training prompt in a chain-of-thought (CoT) manner, to mimic the human reasoning process during fine-tuning.
% This progressive training pipeline allows the model to build foundational skills before adapting to the challenges of real industrial diagrams.

To rigorously evaluate the performance of existing LLMs and our approach in IC footprint labeling, we construct a systematic benchmark that not only captures real-world complexities, such as varying pin counts and package styles, but also aligns closely with the actual distribution of IC footprint types observed in real-world datasets, covering all major package categories. 
The benchmark is detailed in Section~\ref{sec:dataset_analysis} and shown in Appendix~\ref{ssec: benchmark}.
% Detailed description of testing dataset can be found in \tai{Section \ref{sec:dataset_analysis}}.
% Each sample is designed to isolate specific tasks, including pin counting, pin dimension estimation and pin localization.
In addition, to assess the practical utility of our model, we compare its performance against manual footprint creation by experienced electrical engineers using standard EDA tools, focusing on both accuracy and time efficiency. Our main contributions of this article are threefold:
\begin{itemize}
    \item To the best of our knowledge, we are the first to develop a novel benchmark on IC footprint geometry understanding, {\nametest}, and investigate the capability of state-of-the-art general-purpose LLMs (GPT-4o~\cite{hurst2024gpt}, Gemini 2.0~\cite{gemini2_2025}, DeepSeek-VL2~\cite{wu2024deepseek}, and Qwen2-VL~\cite{wang2024qwen2}) on this problem. As shown in Table~\ref{tab:evaluation_overall}, these models struggle with precise geometric reasoning, achieving low overall layout accuracy (e.g., $IoU_{IC}$: GPT-4o 11.1\%, Qwen2-VL 1.7\%) and exhibiting high errors in pin localization and dimension estimation. In contrast, our proposed {\name} achieves significantly superior performance with 71.6\% $IoU_{IC}$, demonstrating its strong capability in accurate and fine-grained IC footprint understanding.

    \item We introduce a new multi-modal geometric reasoning dataset, {\namedata}, containing a total of 8,608 labeled samples—4,138 collected from real-world IC footprint diagrams and 4,470 synthetically generated samples. The dataset contains three sub-tasks, each targeting a key aspect of geometric reasoning: (i) counting the number of pins, (ii) computing the center coordinates of individual pins, and (iii) estimating the dimensions of each pin. To construct the synthetic portion, we propose a novel data augmentation tool that transforms publicly available EDA footprint designs into datasheet-style diagrams with aligned annotations.

    \item We propose a two-stage training framework that enables LLMs to accurately understand IC footprint drawings and automatically generate corresponding footprint designs. 
    Our approach achieves SOTA results on the real-world benchmark compared with existing LLMs.
    By emulating the step-by-step reasoning process of human engineers, our method achieves substantial efficiency gains, generating a complete IC footprint in 0.26 min/sample, representing up to a 58× reduction in time compared to traditional EDA tools, while maintaining comparable accuracy.

\end{itemize}

\section{Related Work}
\label{sec:related}

\paragraph{Manual PCB Component Footprint Generation. }

Traditional PCB component geometry generation is labor-intensive and time-consuming, requiring manual interpretation of datasheets, footprint creation, symbol generation, and signal mapping \cite{ni2020footprint, hundredcomponents}. With hundreds of components in modern designs, this manual process becomes a bottleneck \cite{timeconsuming}, prone to inconsistencies, human error, and outdated libraries due to frequent spec updates \cite{humanmistake, libupdate}. The iterative nature of PCB design further complicates manual updates, delaying time-to-market. These limitations underscore the need for automated and data-driven solutions to improve efficiency and reliability.

\paragraph{Automated IC Footprint Geometry Understanding.}

Existing PCB labeling methods focus on the segmentation or classification of IC footprints~\cite{ni2020footprint,yang2024circuit}. %\cite{ni2020footprint} exploits footprint design rules and the IC file name information to classify footprint types using XGBoost and CNN, contributing to footprint design automation. \cite{yang2024circuit} proposes a standard cell recognition and classification algorithm in IC images by combining a semantic segmentation model with image processing techniques. 
However, these methods do not attempt to understand the geometric information of IC pins, leaving the automated labeling process limited to a level above the individual pins.
Although object detection methods~\cite{yolov5,tan2020efficientdet,zhu2020deformable} may help with counting the number and computing the relative size of the IC pins, they cannot handle footprint diagrams with implicit information omitted, as shown in Figure~\ref{fig:banner}(d). Optical Character Recognition (OCR) methods~\cite{du2024svtrv2,duan2025instructocr} can be used to extract diagram texts from datasheet images, but they cannot understand the physical and geometrical meaning of the numerical labels. 
%\cite{hu2023gptr} proposes a Gestalt-perception Transformer model (GPTR) for the novel diagram object detection. \cite{nurminen2020object} utilizes Yolo and OCR for converting legacy schematic diagrams, such as process and instrumentation diagrams (P\&I diagrams), into computer-understandable forms. 
There are works that utilize object detection and OCR for diagram object detection~\cite{hu2023gptr} or converting legacy schematic diagrams~\cite{nurminen2020object}. However, these methods fail to bridge the gap between the extracted annotations and the implicit geometric knowledge. Therefore, a method capable of performing logical reasoning by interpreting both the footprint diagrams and annotation information is needed for fully automated IC footprint geometry labeling.

\paragraph{LLM for Mathematic.}
Generative general-purpose models such as ChatGPT, DeepSeek, and Qwen2 have demonstrated strong generalization abilities across a wide range of tasks without the need for task-specific fine-tuning \cite{bubeck2023sparks,bai2023qwen,guo2025deepseek}.
To address the problem of visual mathematical geometry reasoning, vision-language models like LLaVA, UniChart, and MathVista use large-scale image-text datasets to develop broad visual reasoning capabilities \cite{lu2023mathvista,masry2023unichart,gao2023g}. 
However, no existing work has explored the capabilities of large language models in understanding IC footprint geometry.
\section{Data Collection}
\label{sec:collection}

\begin{figure}
    \centering
    \includegraphics[width=1\linewidth]{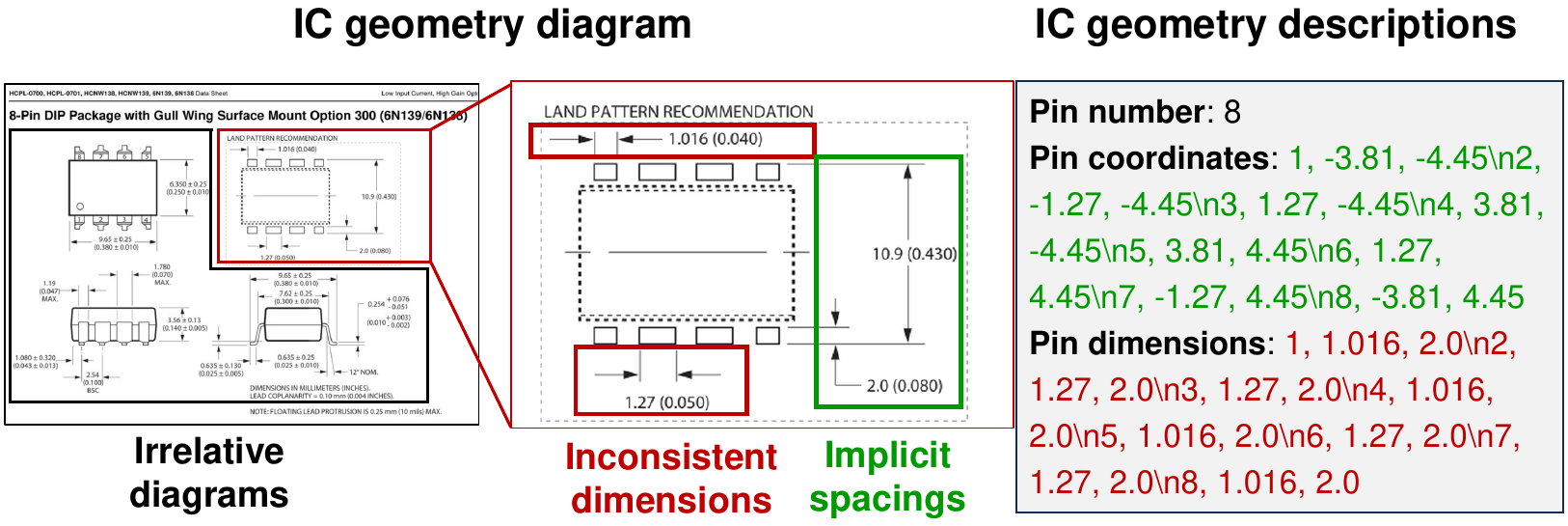}
    \vspace{-0.1in}
		\caption{An example of IC geometry understanding. This starts with locating the correct diagram (upper-right corner). Then the dimensions (framed in \textcolor{red}{red}) and location (framed in \textcolor{green}{green}) of each pin are extracted and computed. Note that datasheet page images often contain irrelevant and misleading information, such as the 3-view diagrams, inconsistent dimension labels, and implicit spacing labels, adding to the complexity.}
		\label{fig:label_example}
        \vspace{-0.2in}
\end{figure}

% Although there exists a large community of PCB engineers and extensive libraries of IC datasheets, there are currently no publicly available datasets specifically curated for describing the pin-level geometry of IC layout diagrams. 
% Although there are many PCB engineers and extensive IC datasheet libraries, there are no publicly available IC footprint datasets specifically focused on pin-level IC layout geometry.
% Traditionally, PCB engineers interpret these datasheet diagrams into EDA files. However, the resulting EDA representations often differ from the original datasheet diagrams, due to engineer's preference, EDA tool-specific constraints and use cases.
% % frequently customize the layouts to suit their particular application requirements. 
% Therefore, we have to collect the IC layout diagrams and the corresponding EDA descriptions from various sources and align them based on IC names. Subsequently, the EDA descriptions must be carefully corrected to ensure accurate correspondence with the original datasheet diagrams. 
Building an IC footprint dataset is challenging due to the lack of a complete index of IC models and the absence of a unified source that provides both datasheets and corresponding EDA footprint files. 
This fragmentation requires collecting and aligning data from multiple platforms, making large-scale data acquisition labor-intensive, error-prone, and difficult to automate.
To build a comprehensive IC footprint dataset, we follow a four-step pipeline by integrating resources from multiple platforms, as no single source provides all the necessary data.
Detailed data collection pipeline is discussed following.
(I) \textbf{Datasheet and EDA Collection.} We compile a comprehensive list of IC models from Digi-Key \cite{digikey2025} and collect their associated datasheets, using IC part numbers as unique identifiers. To supplement missing EDA files, we retrieve standardized footprint and CAD models from UltraLibrarian \cite{ultralibrarian}, aligning them with the datasheets based on these part numbers.
(II) \textbf{AI-Aided Diagram Page Finding.} Because IC footprint diagrams in datasheets are often embedded within extensive technical content, we leverage general-purpose LLMs such as Gemini 2.0 \cite{gemini2_2025} to identify the page numbers containing relevant IC footprint diagrams. Manual verification is subsequently performed to ensure the accuracy and relevance of the extracted images prior to their inclusion in the dataset.
(III) \textbf{Image and Label Processing.} The IC footprint diagrams extracted from datasheets are aligned with the corresponding geometric descriptions from EDA source files using IC part numbers as identifiers. From this alignment, we extract key properties, including the number of pins, their center coordinates, and their dimensions, to construct fully labeled examples for the dataset.
(IV) \textbf{Expert Data Correction.} Expert engineers are engaged to review and correct mismatches between EDA-derived labels and datasheet diagrams, addressing inconsistencies due to design variability.
% This multi-source strategy is necessary because a complete and aligned collection of IC lists, datasheets, and EDA files is not available through any single platform.
An example illustrating IC geometry understanding is presented in Figure~\ref{fig:label_example} and Appendix~\ref{ssec: sample_example}, where the IC geometry diagram is embedded within a datasheet page, and the corresponding geometry descriptions are extracted from the associated EDA file and organized in JSON format.

\subsection{Dataset Building}

% To build a IC footprint dataset, we propose a four-phase dataset building (as shown in Figure~\ref{fig:data_process} ) to construct a specialized IC geometry labeling dataset containing 4138 real-world IC footprint diagrams. The four phases of dataset construction are detailed below.

\paragraph{Phase I: Datasheet and EDA collecting.}
Our dataset consists of paired samples, comprising IC footprint diagrams as input data and corresponding IC geometry descriptions as structured labels (see Figure~\ref{fig:label_example}). 
To construct this dataset, we first extract a broad list of IC models from Digi-Key \cite{digikey2025}, a global distributor offering over 13 million electronic products from more than 2,000 manufacturers. IC part numbers are used as unique identifiers to ensure consistent alignment across different data sources.
We then collect the corresponding datasheets from Digi-Key, which provide essential design information such as footprint diagrams, mechanical package types, and reference application circuits. 
Due to the limited availability of EDA resources on Digi-Key, we augment the dataset by retrieving standardized footprint source files, based on the part number from UltraLibrarian \cite{ultralibrarian}, a widely used platform for manufacturer-approved EDA content.
The IC geometry descriptions, including properties such as pin count, center coordinates, and physical dimensions, are extracted from the EDA source files and aligned with the visual footprint diagrams.
%Detailed dataset collection is discussed in Appendix \ref{ssec: benchmark}.

% \textbf{IC datasheets:} The Digi-Key website is a globally renowned online platform for sourcing electronic components. It assembles over $13$ million IC products from over $2000$ manufacturers and provides access to datasheets, CAD models, and application notes for each product. The datasheet of an IC product describes the features of the component, such as the IC part name, package types, package options, typical applications, and suggested pad layout. 
%We aim to use the suggested pad layout diagrams from datasheets as pin-wise geometry diagrams. 

\paragraph{Phase II: AI-aided diagram page finding.}
% As stated in Phase I, the datasheet of an IC part is a technical PDF file containing various descriptions. Among these descriptions, the \textbf{suggested pad layout} of an IC is a top-view plane figure with clear pin layout visualization and numerical notations, as shown in Figure~\ref{fig:banner}(b), ideal for IC pin recognition and geometric labeling tasks. 

% Extracting the locations of suggested pad layout images from IC datasheets poses several challenges. First, datasheets often contain numerous pages with dense and cluttered content, making manual search inefficient (as illustrated in Figure~\ref{fig:label_example}). Second, automated extraction methods based on simple keyword searches or image pattern recognition face difficulties due to inconsistent presentation formats across datasheets. For instance, terms such as “Recommended layout,” “Land pattern example,” “Footprint recommendation,” and “Suggested PCB layout” are all commonly used to denote suggested pad layouts, complicating keyword-based detection. Additionally, as shown in Figure~\ref{fig:distribution}(a), the wide diversity of IC designs results in significant variations in layout patterns, further hindering effective automated extraction.
Because datasheets use different formats, terms, and visual styles, and often include multiple footprint diagrams for different versions of the same chip, it is difficult to directly use them as input for training without first carefully processing and selecting the correct IC footprint images.
% However, the large number of pages and cluttered content of IC datasheets make it difficult to manually extract the locations of the suggested pad layout images efficiently. As the suggested pad layout images often contain distinguishable features such as entitled ``Recommended layout'' or ``Land pattern example'', we utilize Gemini 2.0 to assist us in finding the PDF pages where the suggested pad layouts are located. 
To address this challenge, we utilize Gemini 2.0\cite{gemini2_2025} to assist with identifying the datasheet pages that contain IC footprint diagrams. 
Given that the model’s predictions are not always precise, manual verification and correction by expert engineers are conducted to ensure accurate localization. The verified pages containing suggested pad layout diagrams are then extracted and used as image inputs for downstream IC geometry understanding tasks.

\begin{figure}
    \centering
    \includegraphics[width=1\linewidth]{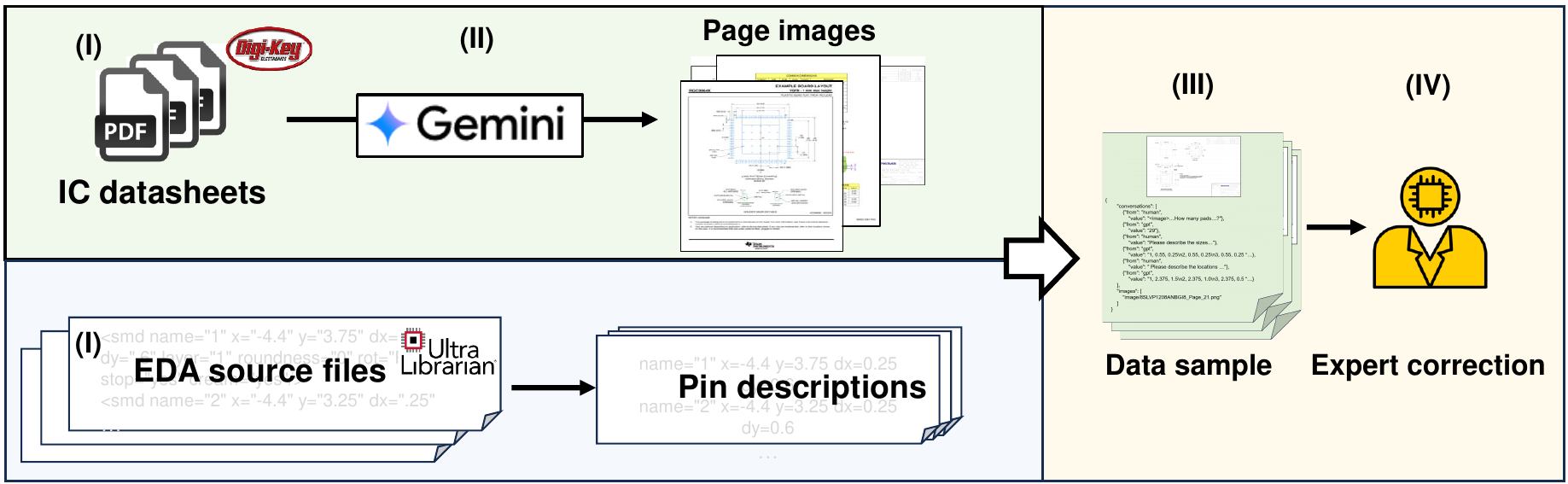}
        \vspace{-0.1in}
		\caption{Dataset building process. A valid sample in {\namedata} is developed in four steps: (I) datasheets and EDA files collection, (II) AI-aided diagram page finding, (III) image and label processing, and (IV) label correction by PCB experts. }
		\label{fig:data_process}
        \vspace{-0.2in}
\end{figure}

\paragraph{Phase III: Image and label processing.}
% Since the image input and numerical descriptions of an IC originate from two distinct sources, datasheet pages from Digi-Key and EDA descriptions from Ultra Librarian, the corresponding images and labels are aligned based on their manufacturing part names. From the EDA files, we extract IC pin descriptions, comprising three key properties for each IC: \textbf{the number of pins}, \textbf{the center coordinates of the pins}, and \textbf{the pin dimensions}. These properties collectively describe the geometry diagram, forming a dataset entry as illustrated in Figure~\ref{fig:label_example}.
After identifying and verifying the page number containing the IC footprint diagrams in the datasheet, we extract that pages as images to serve as the input data. 
In parallel, we preprocess the corresponding EDA source file to extract structured pin-level information, as they  often include extra, unrelated design details such as general component information, wiring data, and board layout elements. 
To focus only on the footprint geometry, we extract and organize three key properties: (i) the total number of pins, (ii) the center coordinates of each pin, and (iii) the size of each pin. These are then formatted to match the label structure shown in Figure~\ref{fig:label_example}.

\paragraph{Phase IV: Expert data correction.}
% Although the numerical descriptions of IC pin layouts in the EDA file generally correspond to the suggested pad layouts provided in the datasheet for the same IC part, discrepancies often arise between the pin dimensions and locations as described in the EDA file and those specified in the datasheet.  
Although the number of pins and their relative positions in the EDA file are generally consistent with the IC footprint diagrams in the datasheet, discrepancies often arise in the exact pin dimensions and precise locations.
%Detailed examples are discussed in Appendix \ref{ssec:datasheets}. 
% These differences can be attributed to the complexity and variability in the EDA files provided by engineers. 
These inconsistencies, caused by variations in EDA file formatting and design conventions, can prevent LLMs from accurately understanding or inferring the geometric details found in EDA files solely from the visual information in datasheet diagrams.
Therefore, we engage human engineering experts to review and correct any inaccurate labels in the EDA files, ensuring that they precisely match the specifications outlined in the IC geometry diagrams. All PCB engineers involved are paid $0.61\$$ per processed sample.
After manual correction and filtering, we carefully selected $4,138$ IC footprint entries, spanning 10 distinct IC package types, each consisting of datasheet-oriented IC footprint diagrams and their corresponding geometry description labels. We name our real-world collected dataset \namedata.

\subsection{Synthetic Diagram Augmentation}
\label{sec:generate-data}

% Although the original EDA descriptions do not precisely align with the suggested pad layout diagrams, they offer diversity in geometry text labels. Therefore, we developed a footprint diagram generation toolkit to convert EDA descriptions into footprint images with realistic geometric notations, thereby augmenting the footprint geometry labeling dataset. An example generated image is shown in figure .
%Due to inconsistencies between footprint diagram annotations in datasheets and those in publicly available EDA file descriptions, direct use as paired data is not feasible. For instance, pin dimensions in EDA footprint designs cannot be reliably derived from datasheet drawings, which often lack precise geometric details.
Due to the time- and labor-intensive nature of manual labeling, collecting a large-scale real-world dataset is challenging, which in turn limits the effectiveness of model fine-tuning.
To overcome this limitation, we developed a footprint diagram generation toolkit that synthesizes clean, datasheet-style footprint images from imprecise EDA geometry descriptions. This approach enables the construction of a synthetic dataset, wherein the synthetic images serve as inputs and the corresponding EDA footprint descriptions provide accurate ground truth annotations for training.
Comparisons of synthetic and real-world diagrams are shown in Figure~\ref{fig:synthetic} in Appendix~\ref{ssec:datasheets}. 
% We generate footprint diagrams from 4,470 selected EDA entries and construct a dataset named \namedatagenerated. 
% Together with \namedata, our final datasets consist of 8,608 entries covering 10 IC types.
%We generate footprint diagrams from 4,470 selected EDA footprint designs to construct a synthetic dataset, \namedatagenerated. 
The synthetic dataset contains 4,470 data samples synthesized from EDA files collected from \textbf{Phase I}. Combined with the real-world dataset with $4,138$ manually labeled real-world IC footprint diagram-label pairs, our final dataset {\namedata} comprises a total of 8,608 entries.
The difference between the synthetic part and the real-world part in entry numbers arises from manual filtering in real-world samples, where real-world diagrams are removed if their pins are vague or irregular.

\section{Training Pipeline}
\label{sec:method}

\begin{figure}
    \centering
    \includegraphics[width=0.9\linewidth]{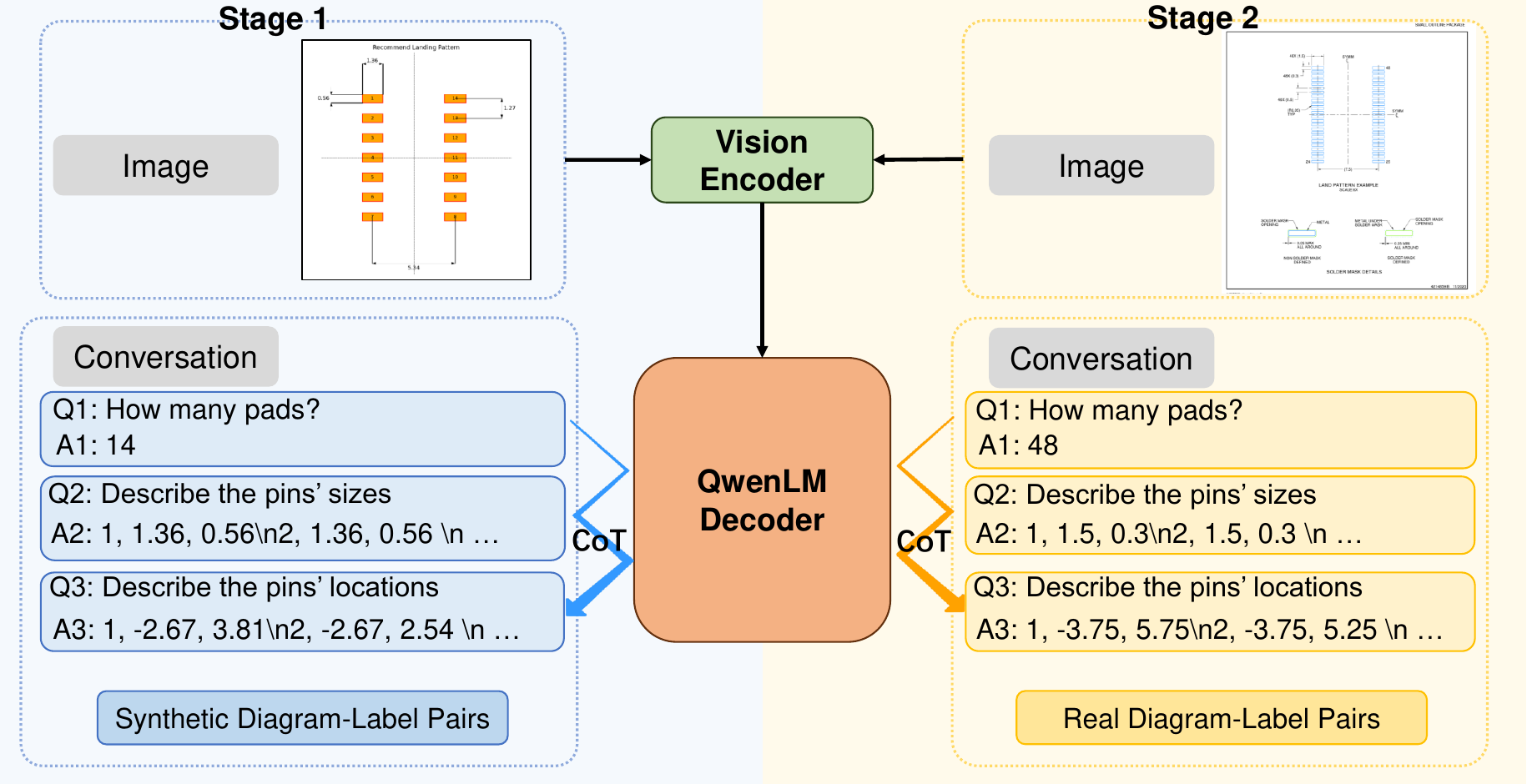}
    \vspace{-0.1in}
		\caption{Two-stage training overview. We apply a two-stage supervised fine-tuning on Qwen2-VL-7B. The questions in a conversation are ordered in a chain-of-thought (CoT) manner. }
		\label{fig:train_stage}
        \vspace{-0.2in}
\end{figure}

To realize the geometric reasoning ability for understanding IC footprint diagrams, this section introduces our proposed {\name} in detail. The main intuition is to leverage the capabilities of LLMs to perceive the relationships between image patterns and annotations, and to derive effective information progressively, similar to how humans process and reason through such information. To achieve this goal, we propose a two-stage, end-to-end IC geometry labeling framework (Figure~\ref{fig:train_stage}) that progressively extracts pin geometries through a chain-of-thought manner.

\subsection{Supervised Fine-tuning with Chain-of-thought}
\label{multi-round}

When human engineers draw EDA footprint based on the IC footprint diagrams in IC datasheets, several critical pieces of information need to be extracted from the diagrams, including the type of IC, the total number of pins, the locations of the pins, and the dimensions of the pins. The thinking process in this procedure follows a progressive approach from the whole to the details: first, the engineer identifies the suggested pin layout diagram in the datasheet and determines the type of IC. Next, the engineer counts the number of pins and identifies their indices. Then, the engineer identifies and reads the spacings between the pins for positioning, followed by determining the pins' widths and lengths to define their dimensions.
%A detailed example illustrating this step-by-step thinking process is provided in Appendix~\ref{ssec: sample_example}.
Drawing inspiration from this reasoning approach, {\name} should also mimic this thinking process to derive the geometric information in a step-by-step manner.

To imitate human reasoning logic, we structure each query conversation for an IC geometry labeling task in three progressive questions: (1) \emph{the number of pins in the IC footprint diagram}, (2) \emph{the coordinates of each pin relative to the center of the diagram}, and (3) \emph{the dimensions of each pin in millimeters}. \textbf{Question 1} directs the LLM to locate the correct diagram and identify the pins within the diagram. \textbf{Question 2} guides the LLM to identify each pin in index order and learn how to locate and interpret the corresponding annotations that reflect pin spacings. \textbf{Question 3} guides the LLM to learn how to identify and interpret the corresponding annotations that reflect pin dimensions. Example query conversations are shown in Figure~\ref{fig:train_stage}.

\subsection{Two-stage Training}
\label{two-stage}
To fully fine-tune the model for IC geometry understanding, we propose a two-stage, end-to-end training process. 
The LLM is initially trained on the synthetic samples of {\namedata}, introduced in Section~\ref{sec:generate-data}. Since the IC diagrams in the synthetic part are generated from EDA descriptions using a plotting toolkit, the annotation rules and graphic patterns are relatively simple and uniform. This allows the LLM to acquire basic geometric reasoning capabilities without interference from complex real-world distractions.
After the first round of training, the LLM undergoes further training on the real-world samples of {\namedata}, where the target diagrams are embedded within datasheet pages. These real-world diagrams vary in image resolution, annotation rules, label text fonts, and presentation styles (examples shown in Figure~\ref{fig:type_example} in Appendix~\ref{ssec: benchmark}). Additionally, multiple diagrams may appear on the same datasheet page, increasing the complexity of identifying the desired diagram. During the second training stage, the model adapts to complex real-world scenarios, enhancing its ability to understand geometric information in IC footprint diagrams.

\section{Experiments}
\label{sec:experiments}

\subsection{Dataset Analysis}
\label{sec:dataset_analysis}

We utilize proposed {\namedata} dataset to train the LLM. 
%Since both {\namedata} and {\namedatagenerated} contain the same list of ICs, the distributions of the two datasets are identical. Therefore, we focus on the analysis of {\namedata}.
The package type of ICs can be categorized into $10$ categories based on their pin types and placements, based on established standards and guidelines provided in \cite{ipc7351b,tiPackagingGuide}. 
Our dataset covers all package categories, reflecting the real-world distribution of common IC footprints. {\namedata} possesses IC footprints with pin counts ranging from $1$ to $800$, implying the complexity in scales for IC pin labeling.
% \begin{figure}
%     \centering
%     \subfigure[]{
%     \includegraphics[width=0.27\linewidth]{figures/type_example.pdf}
%     }
%     \subfigure[]{
%     \includegraphics[width=0.33\linewidth]{figures/distribution.pdf}
%     }
%     \subfigure[]{
%     \includegraphics[width=0.33\linewidth]{figures/pin_count_histograms_test.pdf}
%     }
%     \vspace{-0.1in}
% 		\caption{Benchmark data distribution. (a) Examples of different diagram patterns for distinct package types. (b) The package type distribution of {\nametest} follows that of the 200K IC footprint collection. (c) The pin count distribution of {\nametest}, indicating the diversity in pin count cases.  }
% 		\label{fig:distribution}
%         \vspace{-0.2in}
% \end{figure}

As no prior work addresses pin-level IC geometry labeling, we introduce a novel benchmark, \textbf{\nametest}, to systematically evaluate the performance of LLMs in understanding IC footprint geometry. {\nametest} consists of $400$ carefully curated real-world IC entries sampled from {\namedata}. 
To ensure that the benchmark reliably reflects real-world IC diversity, we analyzed the package type distribution across $200,000$ IC entries crawled from Digi-Key. 
The analysis confirms that {\nametest} exhibits an identical package category distribution to the full 200K dataset, and it shares the same pin count distribution as {\namedata}. 
These validations demonstrate that {\nametest} can reflect the distribution of real-world IC geometries and satisfy the needs of PCB engineers.
% These statistical features ensure that {\nametest} can reflect the distribution of real-world IC geometries and satisfy the needs of PCB engineers. 
Note that the IC entries in {\nametest} are excluded from {\namedata} during the training process. 
% For detailed statistical descriptions of our datasets, please refer to the Supplementary Material.
Please refer to Appendix~\ref{ssec: benchmark} and the Supplementary Material for further detailed descriptions of our benchmark and datasets.

\subsection{Experiment Setup}
\label{ssec: setup}
\paragraph{Implementation Details.} We fine-tune a footprint geometry understanding model based on Qwen2-VL~\cite{wang2024qwen2}, a SOTA LLM in image understanding with Naive Dynamic Resolution mapping and Multimodal Rotary Position Embedding (M-ROPE). As a balance between performance and computational cost, we choose its 7B version (Qwen2-VL-7B) as our base model. We utilize LLaMA-Factory~\cite{zheng2024llamafactory} to fine-tune our model. As stated in Section~\ref{two-stage}, we first implement SFT on the LLM with the synthetic training samples for $3$ epochs, and then fine-tune the LLM with real-world samples for $3$ epochs. We use Low-Rank Adaptation (LoRA) for model training and set the cut-off length to $4096$ and the learning rate to $5e^{-5}$. During each training stage, we randomly split $10\%$ training samples as a validation set. All experiments are conducted on $2$ NVIDIA A100-40G GPUs, and the batch size is set to $2$ per GPU at both training stages. 

\paragraph{Evaluation Metric.}
As no existing work focuses on pin-level IC geometry labeling, we develop our own metric, $IoU_{IC}$, defined as:
\begin{equation}
    IoU_{IC}=\frac{Area_{pred} \cap Area_{label}}{Area_{pred} \cup Area_{label}}
\end{equation}
where $Area_{pred}$ is the area of the pin layout reconstructed from the predicted pin geometry, and $Area_{label}$ is the area of the pin layout reconstructed from ground truth pin geometry. $IoU_{IC}$ ranges from $0$ to $1$, where a value of $1$ indicates the predicted pin layout completely overlaps with the ground truth pin layout, meaning the LLMs provides the correct answer precisely. 
%
% A value of $0$, on the other hand, indicates that the predicted and ground truth layouts do not overlap at all, meaning the answer is entirely incorrect.
%
A value of $0$ indicates no overlap between predicted and ground truth layouts, denoting a completely incorrect result.

As stated in Section~\ref{multi-round}, the IC pin geometry is described by three sub-questions: the number of pins, the position of each pin, and the dimension of each pin. Therefore, we also evaluate model performance for each sub-task. For \textbf{task 1 (pin counting)}, we use Mean Absolute Error (MAE) and Root Mean Square Error (RMSE), where lower values indicate a more accurate counting outcome. For \textbf{task 2 (pin positions)}, we calculate the Euclidean distance between each predicted pin's center coordinates and its corresponding ground truth. Then, the distances of all pins are averaged as $d_{pin}$, where a lower value indicates a more accurate position prediction. For \textbf{task 3 (pin dimensions)}, we use the average IoU between the predicted and ground truth pattern of each pin, $IoU_{pin}$, as the evaluation metric. A more accurate dimension prediction achieves an $IoU_{pin}$ closer to $1$.

\subsection{Comparison with General LLMs}
To evaluate the IC geometric reasoning capability of {\name}, we make comprehensive comparisons between current SOTA general LLMs, namely GPT-4o~\cite{hurst2024gpt}, Gemini 2.0~\cite{gemini2_2025}, DeepSeek-VL2~\cite{wu2024deepseek}, and Qwen2-VL-7B~\cite{wang2024qwen2} with {\name} on {\nametest}. All general LLMs are called via API. To ensure the outputs of general LLMs are compatible and have a consistent format as that of {\name}, we utilize single-shot prompt engineering to inform the LLMs of the output formats. 

% \begin{table}[ht]
% \centering
% \caption{Comparison of QA performance with general LLMs}
% \label{tab:evaluation_overall}
% \begin{tabular}{l c c c c c}
% \hline
% \multirow{3}{*}{\textbf{Methods}} & \multicolumn{4}{c}{\textbf{Metrics}} \\ \cline{2-6}
%                                   & \multirow{2}{*}{Overall ($IoU_{IC}$ (\%))} & \multicolumn{2}{c}{task 1} & \multirow{2}{*}{task 2 ($d_{pin}$)} & \multirow{2}{*}{task 3 ($IoU_{pin}$ (\%))} \\ \cline{3-4}
%                                   &                         & MAE & RMSE &                     \\ \hline
% GPT-4o                          & 11.06                    & 8.2053          & 23.0422           & 4.0102                &45.63\\ 
                                   
% Gemini v2                          & 4.49                    & 1.844          & 7.8785           & 18.2745                &57.34\\ 
                                  
% DeepSeek vl                          & 1.48                    & 21.9735          & 41.6815           & 4.2223                &0.2098\\ 
                                  
% Qwen2-vl-7B                          & 1.69                    & 19.1737          & 43.7584           & 3.6254                &0.4115\\ 

% \textbf{\name}                          & \textbf{71.55}                    & \textbf{0.3467}          & \textbf{2.8088}           & \textbf{1.1115}                &\textbf{87.97}\\ 
% \hline
% \end{tabular}
% \end{table}

\begin{table}[h]
\centering
% \caption{Comparison of QA performance with general LLMs. All results in mean + std format upon 3 individual tests.}
\caption{Comparison of QA performance with general LLMs on 3 tasks in mean + std format}
\vspace{-0.1in}
\label{tab:evaluation_overall}
% \vspace{-0.1in}
\resizebox{\textwidth}{!}{%
\begin{tabular}{l c c c c c}
\hline
\multirow{2}{*}{\textbf{Methods}} & \multirow{2}{*}{\textbf{Overall ($IoU_{IC}$ \%)} } & \multicolumn{2}{c}{\textbf{Task 1}} & \multirow{2}{*}{\textbf{Task 2 ($d_{pin}$)}} & \multirow{2}{*}{\textbf{Task 3 ($IoU_{pin}$ \%)}} \\
\cline{3-4}
& & MAE & RMSE & & \\
\hline
GPT-4o          & $11.1\pm0.4$   & $8.21\pm0.47$   & $23.04\pm0.22$  & $4.01\pm0.02$  & $45.6\pm0.3$ \\
Gemini 2.0       & $4.5\pm0.1$   & $1.84\pm0.40$   & $7.87\pm0.85$   & $18.27\pm0.45$ & $57.3\pm0.4$ \\
DeepSeek-VL     & $1.5\pm0.8$   & $21.97\pm0.35$  & $41.70\pm2.15$  & $4.22\pm0.23$  & $20.1\pm0.2$ \\
Qwen2-VL-7B     & $1.7\pm0.1$   & $19.17\pm0.37$  & $43.76\pm0.84$  & $3.63\pm0.54$  & $41.1\pm0.5$ \\
\textbf{\name}  & \textbf{71.6$\pm$0.5} & \textbf{0.35$\pm$0.07} & \textbf{2.81$\pm$0.08} & \textbf{1.11$\pm$0.02} & \textbf{88.0$\pm$0.3} \\
\hline
\end{tabular}
}
% \vspace{-0.21in}
\end{table}

As shown in Table~\ref{tab:evaluation_overall}, our model demonstrates superior performance among existing SOTA general LLMs. {\name} achieves a $IoU_{IC}$ of $71.6\%$, indicating its ability to label IC geometries accurately, while all general LLMs receive $IoU_{IC}$ below $20\%$, implying failures in labeling IC geometries. Since general LLMs do not have prior knowledge of IC footprint geometry labeling, they fail to understand the meaning of the numerical annotations and graphic symbols in the diagrams, such as mistaking pin spacing annotations as dimensions and miscounting pin numbers with omission symbols. Moreover, general LLMs struggle in accurately identifying numbers, resulting in great misinterpretations of geometric information (Figure~\ref{fig: qualitative results} in Appendix~\ref{ssec: qualitative}). On the other hand, {\name} acquires prior knowledge on IC footprint diagrams, gaining the capability of labeling IC footprint geometry accurately.

\subsection{Ablation Study}\label{sec:experiments_abl}
%To demonstrate the efficacy for our training framework, we conduct ablation studies on various training strategies and factors.

\subsubsection{Different Dialogue Training Strategies}

% As stated in Section~\ref{multi-round}, the problem of IC footprint geometry labeling can be broken into 3 tasks: \textbf{pin number counting (task 1)}, \textbf{pin position understanding (task 2)}, and \textbf{pin dimension understanding (task 3)}. There are differentiations and logistic connections among these 3 tasks. For example, counting pin numbers can be a relatively simple task, but it is impossible to give the complete answers to the latter two questions if the pin number is counted incorrectly at first. Therefore, we evaluate the QA performance under various training strategies concerning different dialogue logics in Table~\ref{tab:evaluation_QA}.
As described in Section~\ref{multi-round}, the problem of IC footprint geometry labeling can be divided into three distinct yet logically connected tasks: \textbf{pin number counting (task 1)}, \textbf{pin position understanding (task 2)}, and \textbf{pin dimension understanding (task 3)}. These tasks differ in complexity and interdependency. For instance, accurately counting pin numbers may be relatively straightforward, yet errors made at this initial stage will prevent correct solutions to the subsequent tasks. Considering these logical relationships, we evaluate the QA performance across various training strategies, reflecting different dialogue sequencing approaches.

We organize our training strategies in each training stage as shown in Table~\ref{tab:combined_strategy}. 
% (detailed in  the Supplementary Material). 
Each training round using different training samples is presented in curly braces (\{\}), the digits $1$, $2$, and $3$ denote the QAs of the corresponding task, and the order of the digits represents the order of the tasks in the conversation. The commas denote that the tasks are separated into independent samples. %For a clearer representation.

% \begin{table}[ht]
% \centering
% \caption{Dialogue training strategy summary}
% \label{tab:QA_summary}
% \begin{tabular}{c c c}
% \hline
% \textbf{Strategy} & {\textbf{Training rounds per stage}} & \textbf{Task order} \\ \hline
% S1 (\name)                          & 1                    & \{123\}\\ 
                                   
% S2                          & 1                    & \{1,2,3\}\\ 
                                  
% S3                          & 2                    & \{1\}\{23\}\\ 
                                  
% S4                          & 2                    & \{12\}\{13\}\\ 

% S5                          & 3                    & \{1\}\{2\}\{3\}\\ 
% \hline
% \end{tabular}
% \end{table}

\begin{table}
\setlength{\abovecaptionskip}{0.1in}
\begin{minipage}{0.65\linewidth}
\centering
\caption{Summary of dialogue and dataset training strategies}
%\vspace{-0.1in}
\label{tab:combined_strategy}
\resizebox{\textwidth}{!}{%
\begin{tabular}{c c c | c c c}
\hline
\multicolumn{3}{c|}{\textbf{Dialogue Training Strategy}} & \multicolumn{3}{c}{\textbf{Dataset Training Strategy}} \\
\textbf{Strategy} & \textbf{Rounds} & \textbf{Task Order} & \textbf{Strategy} & \textbf{Stages} & \textbf{Data Order} \\
\hline
S1 & 1 & \{123\} & T1 & 1 & \{{real-world}\} \\
S2            & 1 & \{1,2,3\} & T2 & 1 & \{real-world, synthetic\} \\
S3            & 2 & \{1\}\{23\} & T3 & 2 & \{real-world\}\{synthetic\} \\
S4            & 2 & \{12\}\{13\} & T4 & 2 & \{synthetic\}\{real-world\} \\
S5            & 3 & \{1\}\{2\}\{3\} & -- & -- & -- \\
\hline
\end{tabular}
}
\end{minipage}
\hspace{4pt}
\begin{minipage}{0.4\linewidth}
\caption{Comparison of performance with existing EDA tools}
\vspace{-0.1in}
\label{tab:label_time}
\resizebox{\textwidth}{!}{%
\begin{tabular}{l c c}
\hline
\textbf{Methods} & \textbf{Overall ($IoU_{IC}$ \%)} & \shortstack{\\\textbf{Time Cost}\\\textbf{(min/sample)}} \\
\hline
Altium           & $95 \pm 0.4$ & 7 \\
EAGLE   & $80 \pm 1.5$  & 15 \\
KiCAD            & $83 \pm 1.7$  & 15 \\
\textbf{\name} & $71.6\pm0.5$  & \textbf{0.26} \\
\hline
\end{tabular}
}
\end{minipage}

\vspace{-0.1in}
\end{table}

% \begin{table}[ht]
% \centering
% \caption{Comparison of performance with existing EDA tools}
% \label{tab:label_time}
% \resizebox{0.6\textwidth}{!}{%
% \begin{tabular}{l c c}
% \hline
% \textbf{Methods} & Overall ($IoU_{IC}$ \%) & \shortstack{\\Time Cost\\(min/sample)} \\
% \hline
% Altium           & $95 \pm 0.4$ & 7 \\
% Autodesk EAGLE   & $80 \pm 1.5$  & 15 \\
% KiCAD            & $83 \pm 1.7$  & 15 \\
% \textbf{LLM4-IC4K} & \textbf{$71.6\pm0.5$}  & \textbf{0.26} \\
% \hline
% \end{tabular}
% }
% \end{table}

\begin{table}[ht]
\centering
\caption{Comparison of QA performance with different training strategies}
\vspace{-0.1in}
\resizebox{\textwidth}{!}{%
\begin{tabular}{c c c c c c c}
\hline
\multirow{3}{*}{\textbf{Factor}} & \multirow{3}{*}{\textbf{Strategy}} & \multicolumn{5}{c}{\textbf{Metrics}} \\ \cline{3-7}
& & \multirow{2}{*}{\textbf{Overall ($IoU_{IC}$ \%)}} & \multicolumn{2}{c}{\textbf{Task 1}} & \multirow{2}{*}{\textbf{Task 2 ($d_{pin}$)}} & \multirow{2}{*}{\textbf{Task 3 ($IoU_{pin}$ \%)}} \\ \cline{4-5}
& & & MAE & RMSE & & \\ \hline

\multirow{5}{*}{Dialogues} 
& \textbf{S1 ({\name})} & \textbf{71.6$\pm$0.5} & $0.35\pm0.07$ & $2.81\pm0.08$ & \textbf{1.11$\pm$0.02} & \textbf{88.0$\pm$0.3} \\
& S2 & $63.5\pm0.4$ & \textbf{0.09$\pm$0.05} & \textbf{0.39$\pm$0.17} & $1.20\pm0.10$ & $82.5\pm0.1$ \\
& S3 & $62.6\pm0.4$ & $5.60\pm0.04$ & $0.63\pm0.14$ & $1.26\pm0.12$ & $85.6\pm0.1$ \\
& S4 & $31.3\pm0.5$ & $0.63\pm0.07$ & $5.67\pm0.04$ & $2.23\pm0.14$ & $75.6\pm0.1$ \\
& S5 & $25.4\pm0.3$ & $10.27\pm0.06$ & $1.09\pm0.21$ & $2.94\pm0.16$ & $74.6\pm0.2$ \\ \hline

\multirow{4}{*}{Training Stages} 
& T1 & $65.1\pm0.1$ & $0.59\pm0.06$ & $3.30\pm0.11$ & $1.17\pm0.03$ & $81.8\pm0.4$ \\
& T2 & $68.2\pm0.3$ & $0.41\pm0.05$ & $2.91\pm0.03$ & \textbf{1.03$\pm$0.06} & $85.3\pm0.3$ \\
& T3 & $24.7\pm0.5$ & $3.17\pm0.06$ & $36.99\pm0.33$ & $3.69\pm0.20$ & $64.9\pm0.2$ \\
& \textbf{T4 ({\name})} & \textbf{71.6$\pm$0.5} & \textbf{0.35$\pm$0.07} & \textbf{2.81$\pm$0.08} & \textbf{$1.11\pm0.02$} & \textbf{88.0$\pm$0.3} \\
\hline
\end{tabular}%
}
\label{tab:evaluation_QA}
% \vspace{-0.3in}
\end{table}

As shown in Table~\ref{tab:evaluation_QA}, the training strategy S1 achieves the highest $IoU_{IC}$ of $71.6\%$. Compared with independent training in S2, the $IoU_{IC}$ of S1 increases by $12.8\%$. This observation suggests that building a chain of thoughts among labeling tasks significantly enhances IC geometry understanding. The sub-optimality of multi-round training leads to biased learning, where the model retains more recent task knowledge more effectively, while earlier knowledge gradually fades.

\subsubsection{The Benefits of the Synthetic Image Dataset}

As stated in Section~\ref{two-stage}, {\name} applies a two-stage training where it is trained under the synthetic part of {\namedata} in the first stage and the real-world part in the second stage. To explore the role of synthetic data in enhancing model performance, we evaluate the QA performance under various training strategies under different dataset part combinations as shown in Table~\ref{tab:evaluation_QA}.
T4 significantly improves performance, increasing $IoU_{IC}$ by $10\%$, $5\%$, and $189.9\%$ compared to T1, T2, and T3, respectively. {\name} gains a deeper understanding of real-world IC geometry by first learning general IC footprint patterns from synthetic data, and subsequently adapting to more complex real-world diagrams.

\subsection{Comparison with Manual Labeling}
To emphasize {\name}'s efficiency on IC footprint geometry labeling, we compare the effort of automated labeling using {\name} with traditional manual labeling using EDA tools. We conduct labeling experiments on {\nametest} with professional IC engineers using three common EDA software: Altium~\cite{altiumIPCWizard}, Autodesk EAGLE~\cite{autodeskEagleLibrary2020}, and KiCAD~\cite{KiCadFootprint}. The labeling times are recorded and averaged by the total sample number. We also record the time of automated labeling using {\name}.
As shown in Table~\ref{tab:label_time}, Autodesk EAGLE and KiCAD require tens of minutes to process an IC diagram. While Altium provides IC templates to accelerate manual labeling, it still costs $7$ minutes on average. On the other hand, automated labeling via {\name} costs only $15$ seconds per image with an accuracy of over $70\%$ in IoU, indicating the groundbreaking evolution our model brings about.

\section{Conclusion and Discussion}
\label{sec:conclusion}

We present the first systematic investigation of general-purpose LLMs in understanding IC footprint geometry, revealing their limitations in precise spatial reasoning tasks. To address this, we introduce {\namedata}, a multi-modal dataset targeting IC geometric understanding, and augment it with {\namedatagenerated} for broader coverage. We develop {\name}, enabling LLMs to accurately interpret footprint diagrams and generate designs with SOTA performance.

While we use a two-stage supervised training to fine-tune the LLM, we are aware that Reinforcement Learning (RL) methods such as Group Relative Policy Optimization (GRPO)~\cite{shao2024deepseekmath} or systematic designs such as the use of agents may further improve model performance. We plan to fully exploit the potential of RL methods and the use of agents in our future work.
We plan to continue scaling up the real-world dataset to establish a more comprehensive and representative benchmark. 
Additionally, by continuously collecting practical data and feedback, we seek to gain deeper insights into the actual needs of IC engineers.
% thereby facilitating the development and support of additional functionalities tailored to real-world engineering applications.
Nevertheless, our work leverages the capability of general LLMs to comprehend complex IC diagram geometry, thereby paving the way toward fully automated PCB engineering and the advancement of geometry-aware LLM development.

\bibliographystyle{unsrt}
\bibliography{paper}

\newpage
\appendix

\section{Appendices}
\label{sec: appendix}

\subsection{Data sample formation example}
\label{ssec: sample_example}

\begin{figure}[H]
\vspace{-0.1in}
    \centering
    \includegraphics[width=0.7\linewidth]{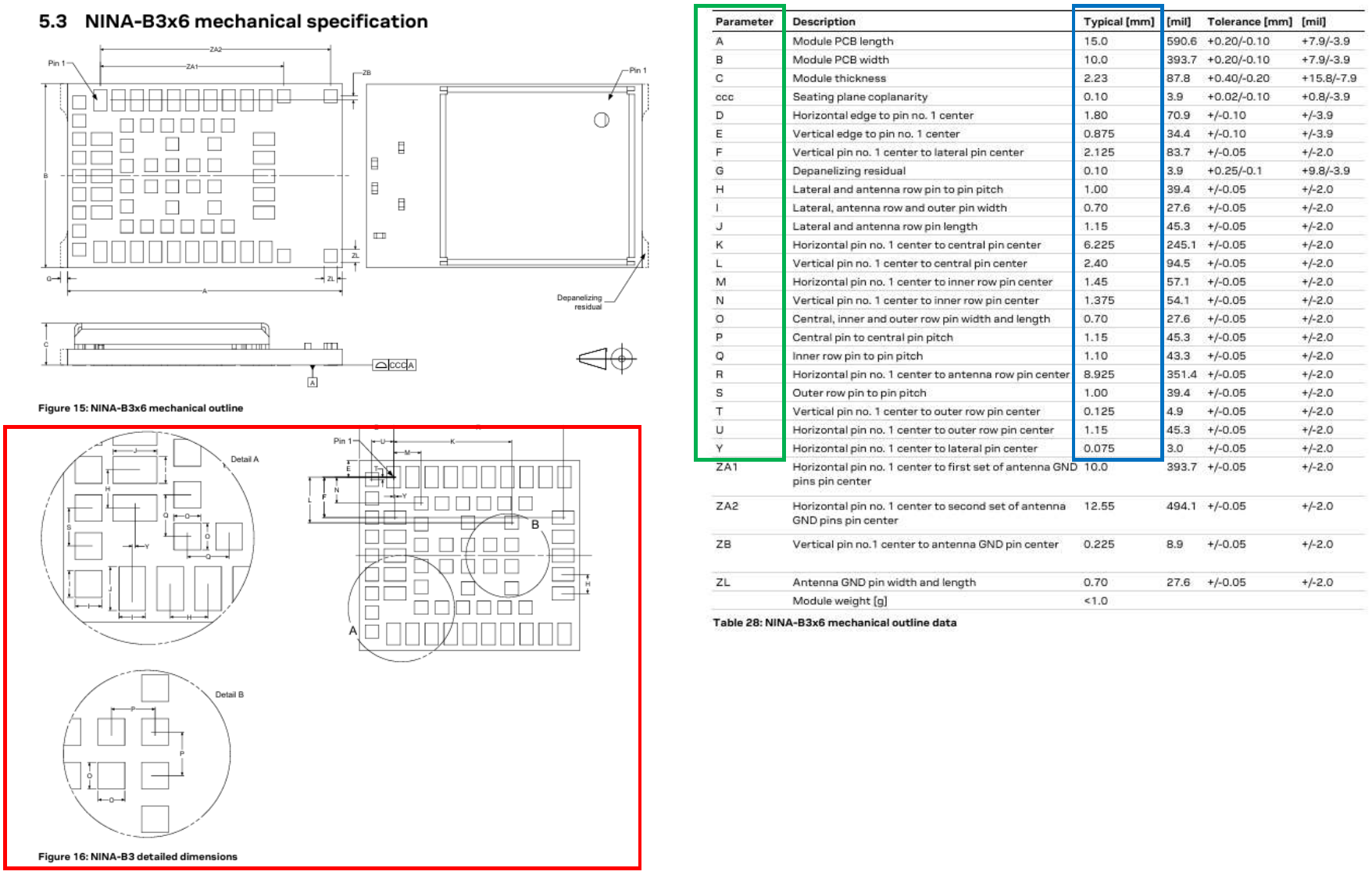}
		\caption{Real-world datasheet example.}
		\label{fig: datasheet_example}
        \vspace{-0.1in}
\end{figure}

This section shows an example of the data processing pipeline for a dataset sample. Figure~\ref{fig: datasheet_example} shows the real-world datasheet pages of the target IC entry. In this example, the information about the IC footprint diagram is distributed over two pages. The two pages are concatenated into one image. The target diagram is located inside the \textcolor{red}{red} frame, while the associated parameter labels about the coordinates and dimensions of pins are framed in \textcolor{green}{green} and the corresponding values are framed in \textcolor{blue}{blue}.

\begin{figure}[H]
\vspace{-0.1in}
    \centering
    \includegraphics[width=0.7\linewidth]{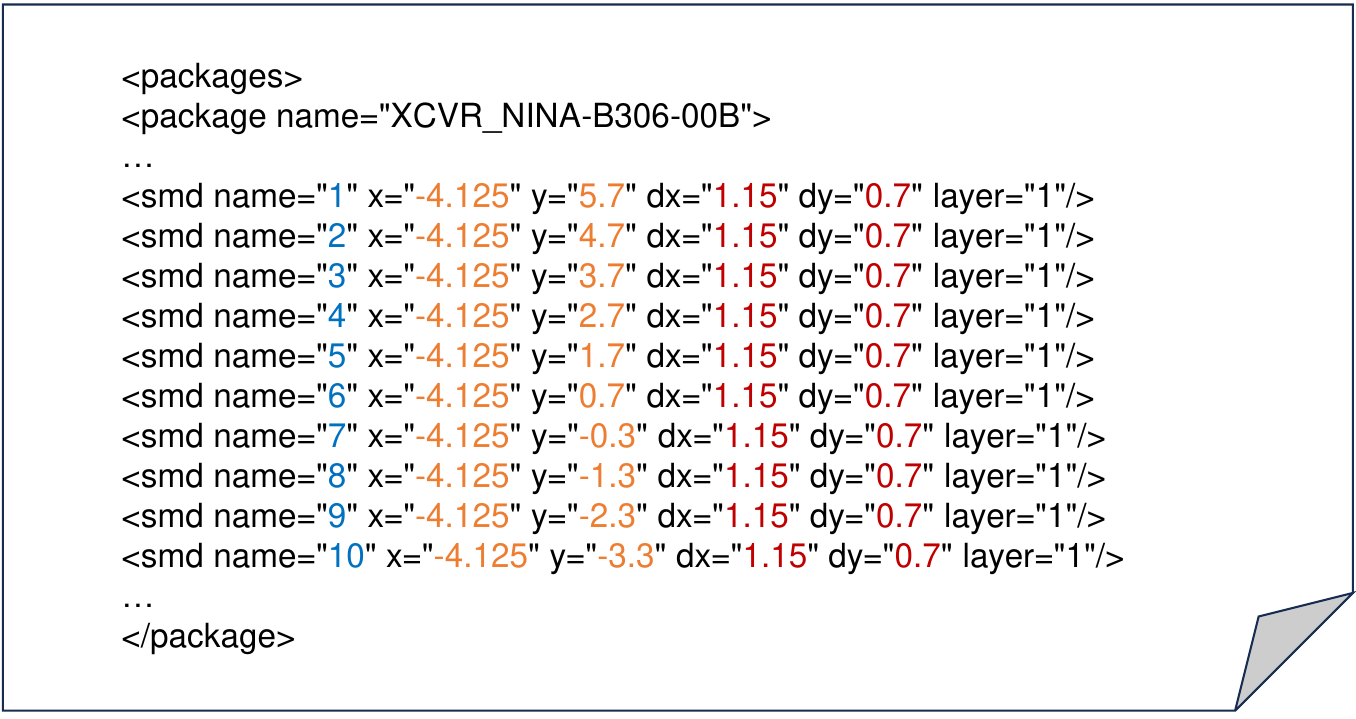}
		\caption{EDA description example.}
		\label{fig: EDA_example}
        \vspace{-0.1in}
\end{figure}

Figure~\ref{fig: EDA_example} shows the EDA description of pins for the same IC entry. The EDA descriptions are in XML format and contain the location and dimension information of each pin. To visualize clearly, the pin indices are colored in \textcolor{blue}{blue}, the coordinates of pins are colored in \textcolor{orange}{orange}, and the dimensions of pins are colored in \textcolor{red}{red}.

\begin{figure}[H]
\vspace{-0.1in}
    \centering
    \includegraphics[width=0.9\linewidth]{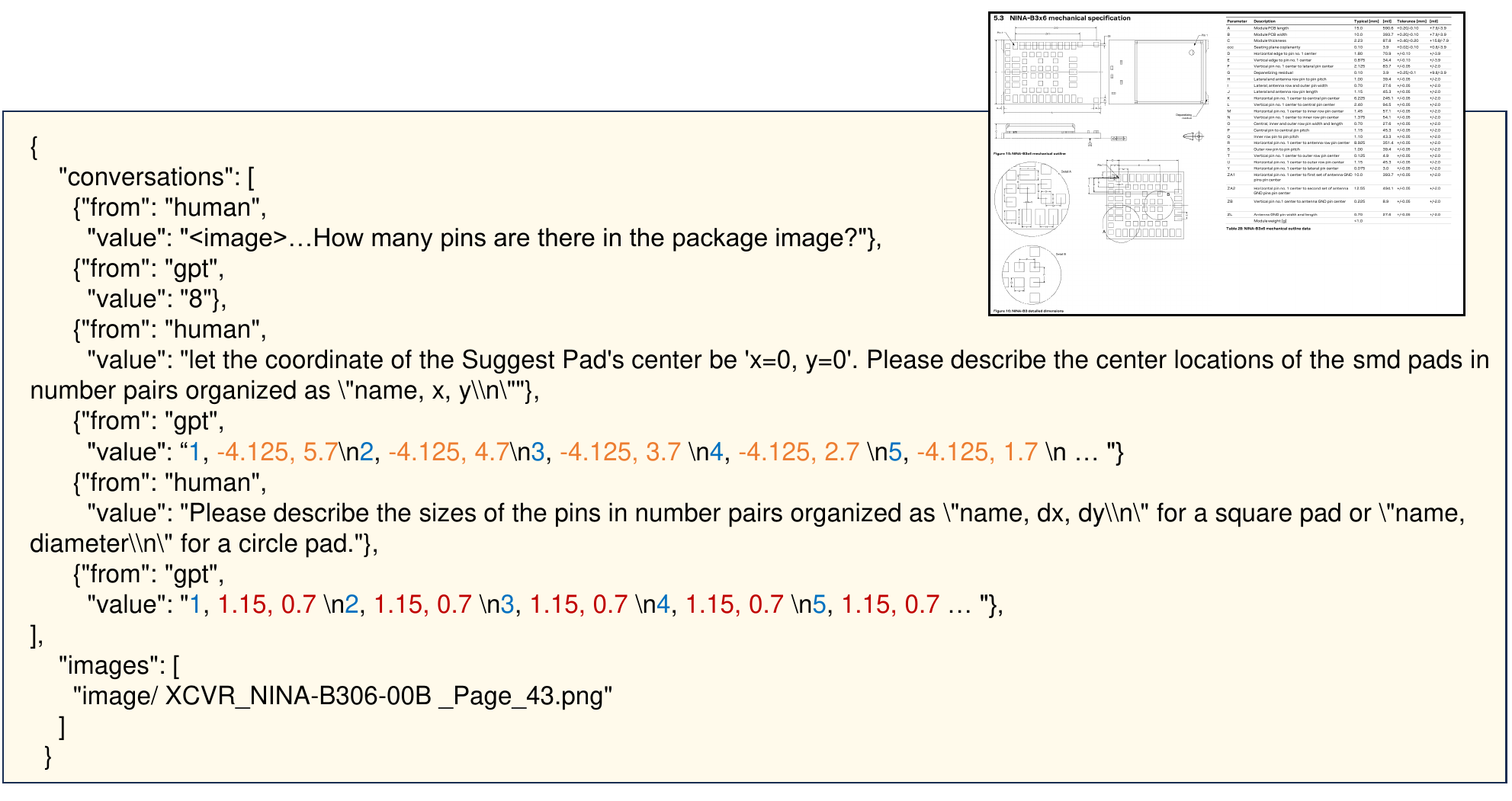}
		\caption{Dataset sample example.}
		\label{fig: sample_example}
        \vspace{-0.1in}
\end{figure}

Figure~\ref{fig: sample_example} shows the final data sample of the IC entry. The data sample comprises the datasheet page image and the JSON format conversation QAs. Note how the values in the three QAs are matched with those in Figure~\ref{fig: EDA_example}.

\subsection{Comparison between synthetic diagrams and real-world datasheets}\label{ssec:datasheets}

Sample 1 and 2 are two distinct ICs. The left-half figure shows the real presentations of the two samples (all real-world diagrams in the following sections are the zoomed-in targeted diagrams), and the right-half figure shows their synthetic counterparts. The synthetic diagrams are simple and unified in colors and label formats, while the real pages contain disturbances and are organized in various formats.

\begin{figure}[H]
\vspace{-0.1in}
    \centering
    \includegraphics[width=1\linewidth]{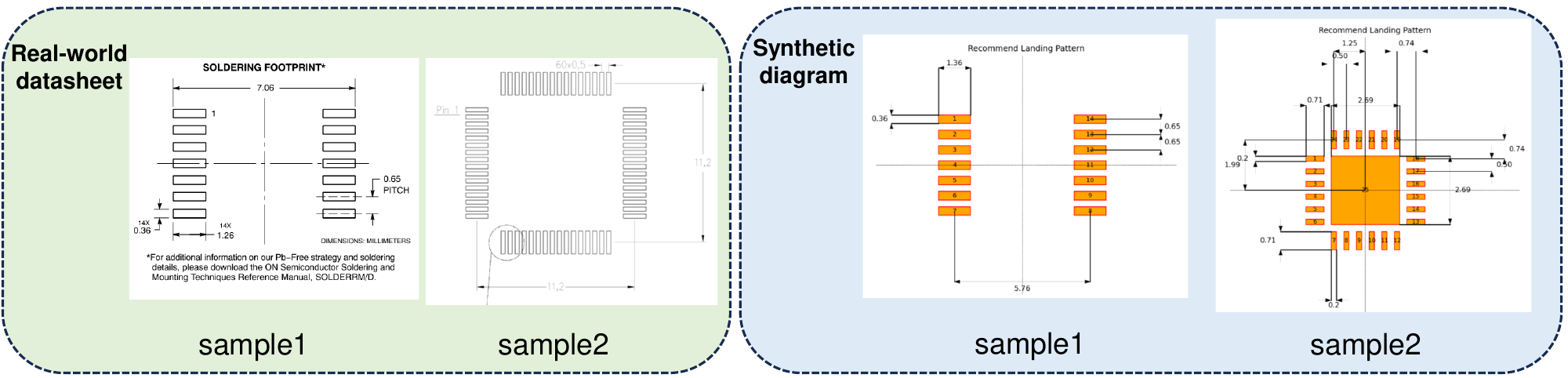}
		\caption{Examples of synthetic diagrams and real-world datasheets. }
		\label{fig:synthetic}
        \vspace{-0.1in}
\end{figure}

\subsection{Benchmark data distribution}
\label{ssec: benchmark}
We downloaded $200K$ IC information entries, of which $50K$ have available datasheets.
\textbf{IC EDA files:} Ultra Librarian is a comprehensive electronic component library offering access to 16 million verified components described in EDA design files. These files provide real-world examples of IC landing patterns, allowing extraction of numerical descriptions of IC parts (\ie, pin positions and sizes). We download EDA design files corresponding to the collected IC datasheets.
%and extract pin layout data as geometry description labels. 
However, since EDA files are individually provided by manufacturers or engineers, over half of the datasheets in the Digi-Key library lack matching EDA files, reducing the pool of valid image-label pairs to fewer than $25K$ entries.

\begin{figure}[H]
\vspace{-0.1in}
    \centering
    \includegraphics[width=0.8\linewidth]{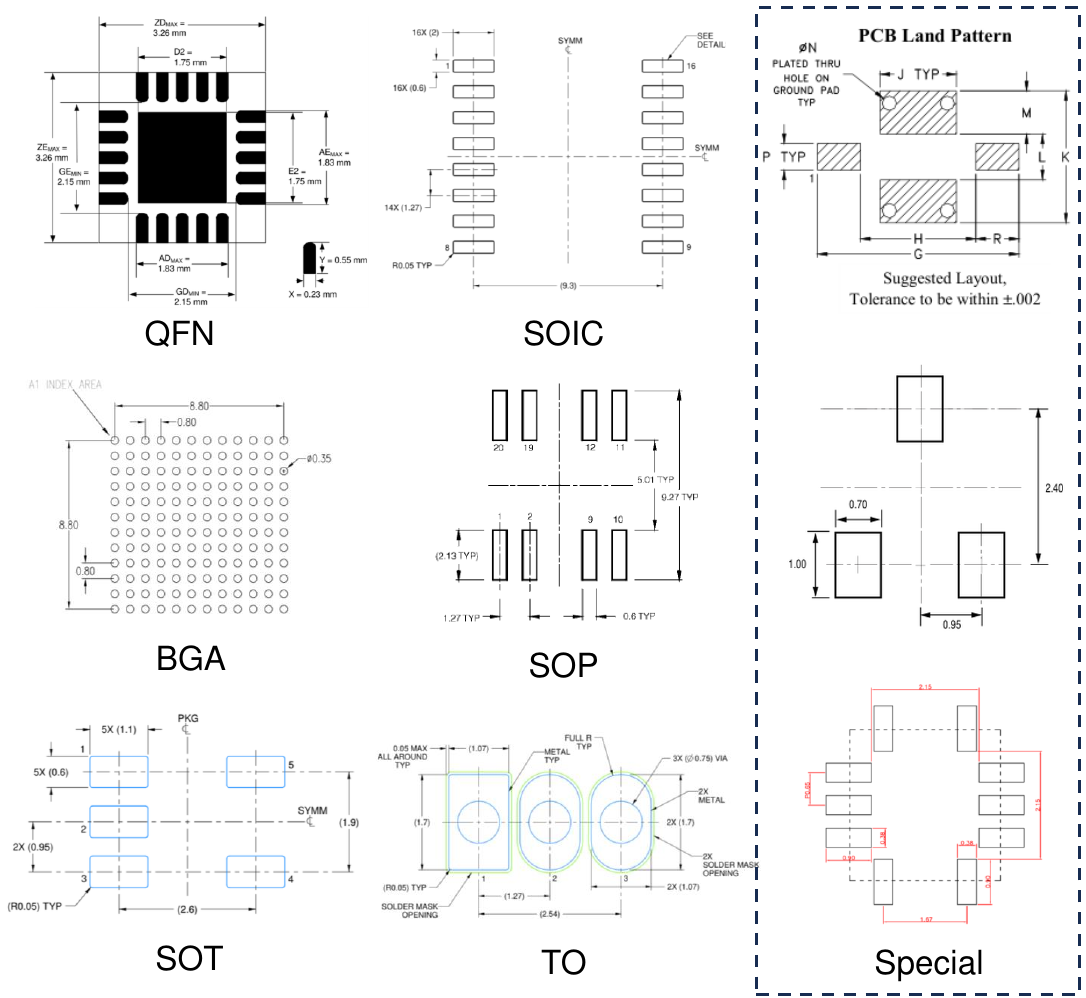}
		\caption{Examples of different diagram patterns for distinct package types. }
		\label{fig:type_example}
        \vspace{-0.1in}
\end{figure}

Figure~\ref{fig:type_example} shows a variety of IC layout patterns for different package types. Different IC package types are largely distinct from each other in pin types, pin numbers, pin shapes, and placements, indicating the complexity of the IC geometry understanding problem.

\begin{figure}[H]
\vspace{-0.1in}
    \centering
    \includegraphics[width=0.7\linewidth]{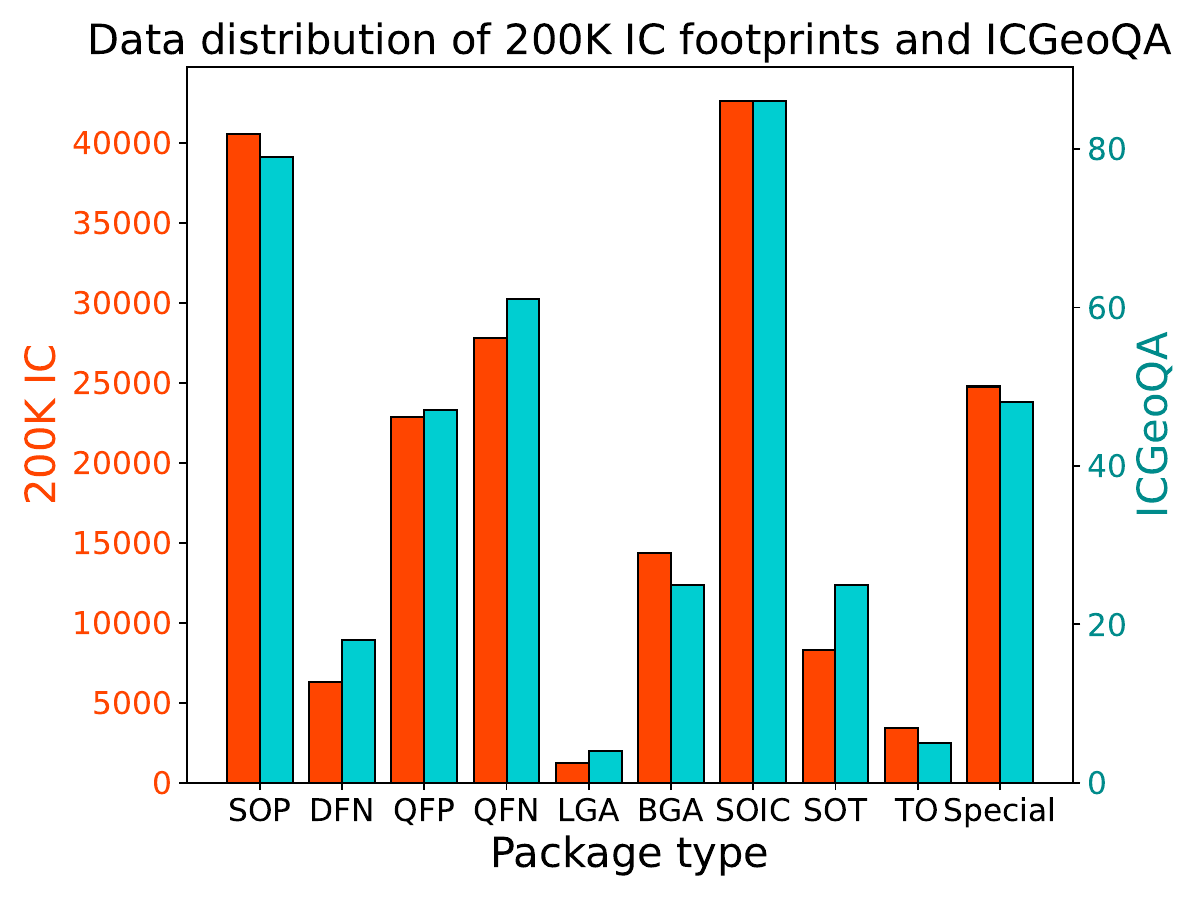}
		\caption{Examples of synthetic diagrams and real-world datasheets. }
		\label{fig:distribution}
        \vspace{-0.1in}
\end{figure}

Figure~\ref{fig:distribution} illustrates the package type distribution of {\nametest} alongside the $200K$ IC footprint entries collected from Digi-Key. The distribution of {\nametest} is deliberately designed to mirror that of the $200K$ IC footprint dataset, ensuring that our benchmark accurately reflects the practical requirements of PCB engineers in their daily design tasks.

\begin{figure}[H]
\vspace{-0.1in}
    \centering
    \includegraphics[width=0.7\linewidth]{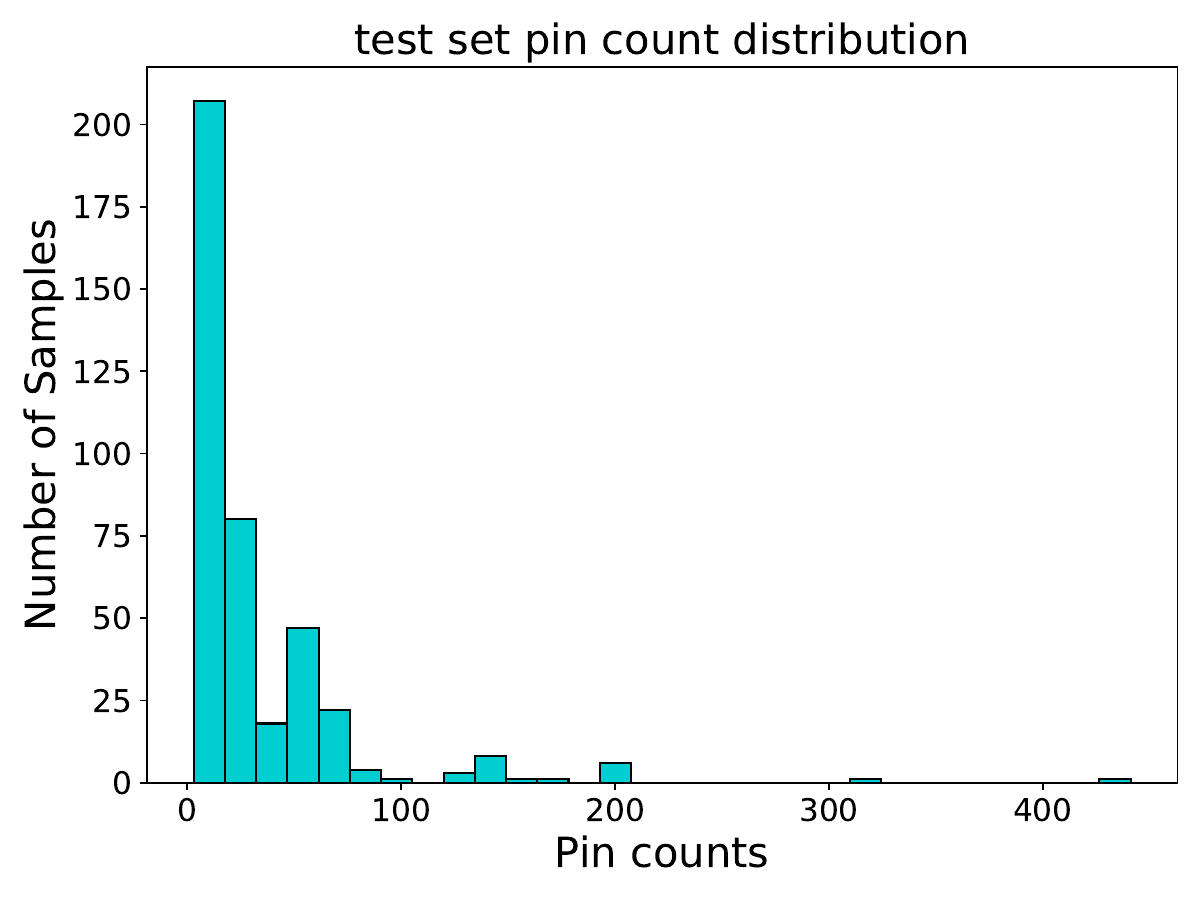}
		\caption{The pin count distribution of {\nametest}. }
		\label{fig:pin_count_histograms_test}
        \vspace{-0.1in}
\end{figure}

Figure~\ref{fig:pin_count_histograms_test} shows the distribution of pin numbers of our benchmark, indicating the diversity in pin count cases. As over $80\%$ of the samples contain less than 100 pins, it's reasonable to set the cut-off length of our model to 4096 to avoid massive token truncation.

\subsection{Qualitative results}
\label{ssec: qualitative}

\begin{figure}[H]
\vspace{-0.1in}
    \centering
    \includegraphics[width=0.8\linewidth]{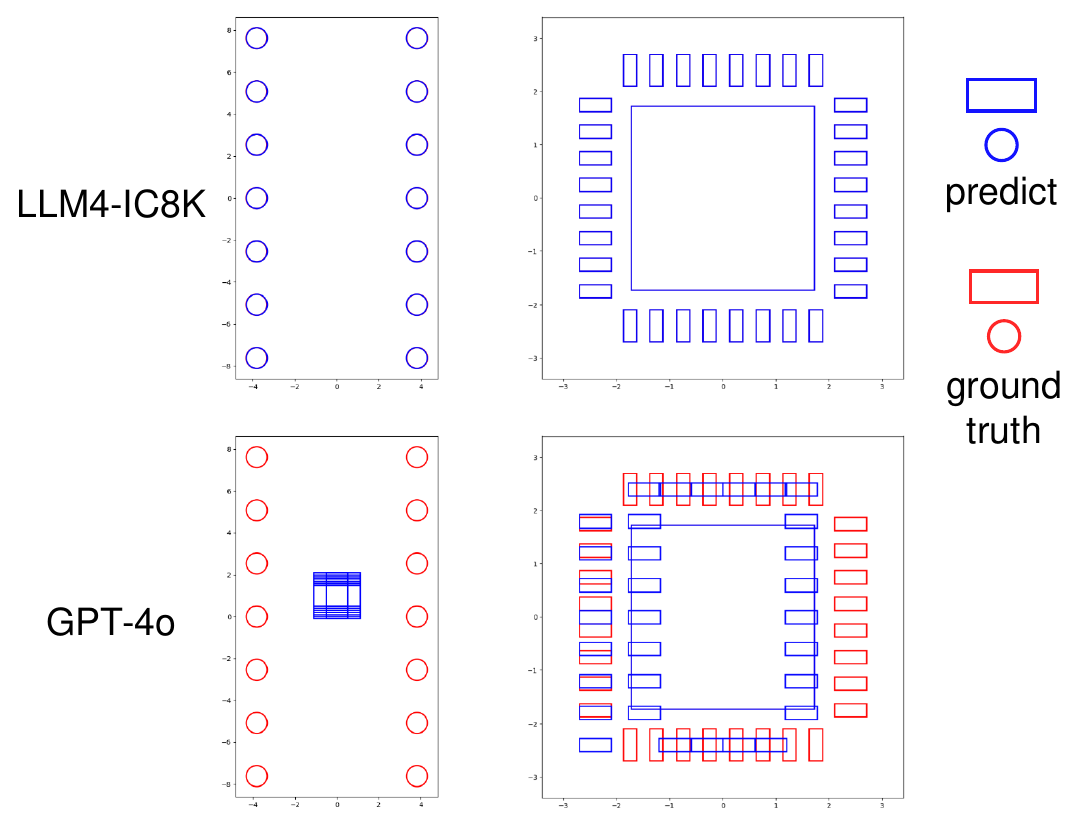}
		\caption{Visualization of pin dimensions and locations of ground truth and generated results.}
		\label{fig: qualitative results}
        \vspace{-0.1in}
\end{figure}

Figure~\ref{fig: qualitative results} shows the qualitative results. In the experiment, {\name} and GPT-4o are both employed to predict the pin dimensions and locations given the same IC diagrams. We then visualized the pin patterns of both ground truth and predicted descriptions. The squares and circles in the images are visualized IC pins according to label descriptions, the predicted results by LLMs are colored in \textcolor{blue}{blue}, and the ground truth images are colored in \textcolor{red}{red}. Note that the results predicted by {\name} overlaps perfectly with the ground truth images, while GPT-4o performances poorly in all 3 sub-tasks. The results suggest that {\name} significantly outperforms the SOTA general LLM (GPT-4o) in predicting IC pin quantities, positions, and dimensions, achieving near-perfect alignment with the ground truth, whereas GPT-4o exhibits significant inconsistencies.

% \newpage
% \input{checklist}

\end{document}